\documentclass[journal]{IEEEtran}
\usepackage{amsmath,amsfonts}
\usepackage{algorithmic}
\usepackage{algorithm}
\usepackage{array}
\usepackage[caption=false,font=normalsize,labelfont=sf,textfont=sf]{subfig}
\usepackage{textcomp}
\usepackage{stfloats}
\usepackage{url}
\usepackage{verbatim}
\usepackage{graphicx}
\usepackage{changepage}
\usepackage{booktabs}
\usepackage{cite}
\usepackage{enumitem}
\usepackage{xcolor}
\newcommand{\rev}[1]{{#1}}

\begin{document}

\title{\rev{REARL: A Closed-loop Autonomous Driving Simulation Enhancement Framework with Real Traffic Data and Large Language Models}}

\author{Xiaojun Bi, Jun Jiang, Yiwen Sun\IEEEauthorrefmark{1}, Quanyi Ou, Yizhi Ma, Ke Cheng, Mingjie Bi, Yexin Li and Bowen Du 
\thanks{This work was supported by the National Natural Science Foundation of China (Grant No. 62503015). (Corresponding author: Yiwen Sun.)}
\thanks{Xiaojun Bi is with the Key Laboratory of Ethnic Language Intelligent Analysis and Security Governance of MOE, Minzu University of China, Beijing, 100081, China(e-mail: bixiaojun@hrbeu.edu.cn).}
\thanks{Yiwen Sun is with the Institute for Artificial Intelligence, Peking University, Beijing, 100871, China, and also with the State Key Laboratory of General Artificial Intelligence, BIGAI, Beijing, 100080, China (e-mail: sunyiwen@pku.edu.cn).}
\thanks{Jun Jiang is with the Key Laboratory of Ethnic Language Intelligent Analysis and Security Governance of MOE, Minzu University of China, Beijing, 100081, China(e-mail: 23302184@muc.edu.cn).}
\thanks{Quanyi Ou is with the Key Laboratory of Ethnic Language Intelligent Analysis and Security Governance of MOE, Minzu University of China, Beijing, 100081, China(e-mail: ouquanyi@muc.edu.cn).}
\thanks{Yizhi Ma is with the Key Laboratory of Ethnic Language Intelligent Analysis and Security Governance of MOE, Minzu University of China, Beijing, 100081, China(e-mail: 25300526@muc.edu.cn).}
\thanks{Ke Cheng is with the School of Computer Science and Engineering, Beihang University, Beijing, 100191, China (e-mail: ckpassenger@buaa.edu.cn).}
\thanks{Mingjie Bi is with State Key Laboratory of General Artificial Intelligence, BIGAI, Beijing, 100080, China (e-mail: bimingjie@bigai.ai).}
\thanks{Yexin Li is with State Key Laboratory of General Artificial Intelligence, BIGAI, Beijing, 100080, China (e-mail: liyexin@bigai.ai).}
\thanks{Bowen Du is with School of Transportation Science and Engineering, Beihang University, Beijing, 100191, China (e-mail: dubowen@buaa.edu.cn).}}

\markboth{Journal of \LaTeX\ Class Files,~Vol.~14, No.~8, August~2021}%
{Shell \MakeLowercase{\textit{et al.}}: A Sample Article Using IEEEtran.cls for IEEE Journals}

\IEEEpubid{0000--0000/00\$00.00~\copyright~2021 IEEE}

\maketitle

\begin{abstract}
Accurate simulation is crucial for autonomous driving development. Yet capturing real-world traffic complexity remains challenging. The existing simulators, which rely on predefined rules or static data playback, struggle to handle dynamic traffic scenarios. To address this, CRITICAL proposes using real traffic data and large language model(LLM) to adjust the initial simulation environment configuration. Based on our experiments, while this improves realism, the data distribution gradually diverges from real-world traffic scenarios as the simulation evolves. We propose a novel framework, \rev{REARL, a closed-loop autonomous driving simulation enhancement framework with real traffic data and large language models}, whose main advantage lies in integrating real traffic data with LLM. First, real-world traffic data employed for enriching the simulation environment is partitioned into multiple categories via a clustering method. The cluster center of each category is defined as a representative scenario, serving as the target simulation object. These representative scenarios constitute typical real-world traffic patterns, which the LLM uses as references to learn the characteristics of real-world traffic distributions. Second, a timed detection mechanism with a sliding window continuously monitors the discrepancies between the current simulation and the most similar representative scenario, specifically in terms of vehicle speed distribution and the mean spacing between pairs of vehicles. Third, if any evaluation metric exceeds a predefined threshold, the LLM intervenes to adjust the vehicle's decision-making; otherwise, the existing control strategy remains unchanged. Finally, based on the current vehicle's driving state, the LLM selects the most relevant vehicle from a real-world traffic data snapshot that best matches the current driving scenario. It then modulates the simulated vehicle's behavior \rev{with reference to the actions of the matched real vehicle}, thereby \rev{promoting more realistic behavior adjustment under the given simulation setting}. Evaluations in the \rev{controlled HighD highway setting} show that REARL achieves \rev{competitive and metric-dependent performance} across the evaluated metrics. Compared to the CRITICAL baseline and a PPO-based learning baseline, it reduces the Hellinger distance for speed distributions to 0.3067, the Mean Absolute Percentage Error (MAPE) for mean spacing between pairs of vehicles to 0.8371, while achieving a higher time headway (THW) of 22.8575, \rev{reflecting more conservative car-following behavior}, and a moderate lane change rate of 0.0708, \rev{suggesting a moderate level of lateral maneuvering under the tested setting}.
\end{abstract}


\section{Introduction}
\label{sec1}
\begin{figure}[htbp]
\centering
\includegraphics[width=0.5\textwidth]{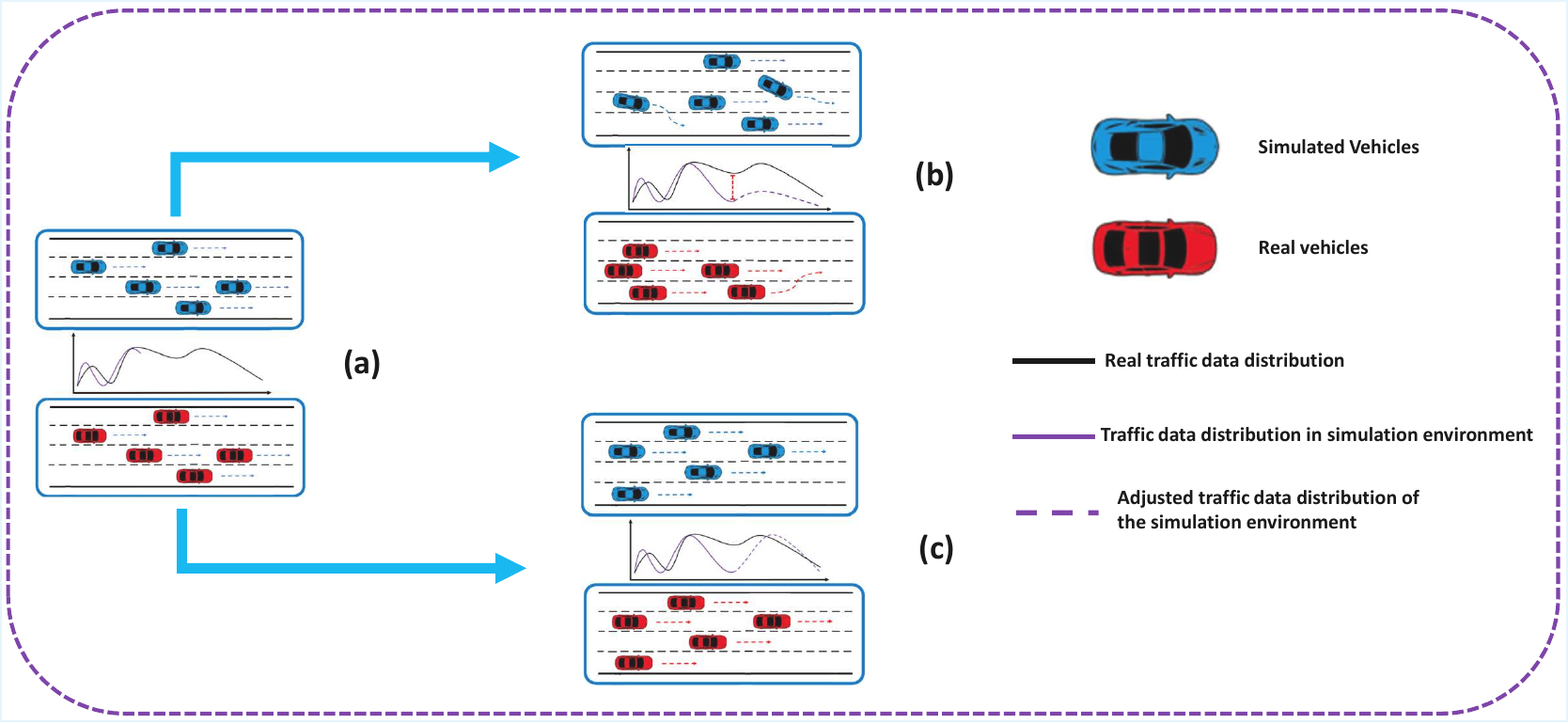}
\caption{In some cases, the simulation environment will gradually deviate from the real data as the process progresses. \rev{The proposed paradigm monitors discrepancies between the simulated vehicle distribution and real traffic data and adjusts the simulation environment during rollout to reduce these discrepancies, as illustrated in Fig. 1(a).}}
\label{fig1}
\end{figure}

\IEEEPARstart{A}{utonomous} driving technology is essential for advancing ITS, enabling efficient and adaptive traffic systems. Simulation environments, as the core platform for testing and validating autonomous driving systems, are becoming increasingly important \cite{3,4}. As a result, enhancing simulation environments is critical for developing autonomous driving. It provides realistic and dynamic testing of algorithms in a controlled virtual setting. Figure \ref{fig1} illustrates the issues faced by autonomous driving simulation environments and their ideal state. Figure 1(a) shows the initial simulation state, where the distribution of simulated vehicles deviates slightly from real traffic data. Figure 1(b) highlights the limitations of methods without simulation environment enhancement: as the simulation environment evolves, it gradually deviates from real data, increasing distortion. Figure 1(c) illustrates the idealized simulation state, where the vehicle speed distribution and the mean spacing between pairs of vehicles in the simulation gradually approach the real-world distribution.

Traditional autonomous driving simulation methods mainly rely on predefined driver models and traffic behavior rules. These include heuristic models or static data based on log playback \cite{11}. These methods work well in simple scenario testing. However, their performance is limited in dynamic environments. They struggle to simulate complex interactions accurately \cite{12}. In high-density traffic scenarios, driver decisions are influenced by factors such as traffic density, mean spacing between pairs of vehicles, and individual driving habits. These factors exhibit significant time-varying and nonlinear characteristics. Traditional models are too simplified. They fail to capture these dynamic features. As a result, simulation results deviate from real-world scenarios.

To address the limitations of traditional methods, data-driven simulation environments have gained increasing attention. These methods integrate and analyze large-scale traffic data. This significantly enhances the dynamic nature and precision of simulations. However, data-driven methods still face challenges such as data bias, overfitting, and poor interpretability. These issues limit their application in complex interaction scenarios \cite{13,14,15,16,17}. Moreover, the randomness and noise characteristics of traffic data make it difficult for purely data-driven models to make reasonable decisions in critical scenarios, especially when responding to emergencies \cite{18,19,20}.

Recent advances in LLMs have opened new possibilities for autonomous driving simulation. For example, \cite{21} propose CRITICAL, using LLMs to optimize the initial simulation configuration, enhancing system intelligence and adaptability. Simulation enhancement technologies, integrating large-scale real data, improve realism by capturing dynamic traffic flow and complex driver behavior. The advantages of LLMs in reasoning and decision-making enable them to simulate traffic behavior based on real-world data \cite{8,18}. However, the application of LLMs in dynamic driving scenarios still faces challenges. These include dependence on historical data and prior knowledge, as well as poor performance in highly unpredictable situations \cite{8,12}. The \rev{CRITICAL} only optimizes the initial simulation configuration, without monitoring and correcting intermediate processes. As a result, it falls short in simulating traffic flow statistical characteristics and flexibly responding to complex interactions.

Current autonomous driving simulation research has the following major shortcomings: First, insufficient simulation of traffic flow statistical characteristics. Existing methods fail to capture the statistical features of dynamic traffic flow and complex interaction scenarios, leading to deviations between simulation and real-world environments \cite{5,6,7,8}. Second, the lack of \rev{closed-loop feedback} mechanisms. Existing rule-based or data-driven simulation systems struggle to simulate dynamic driving behavior in high-density traffic or complex interaction scenarios. Most simulation systems cannot adjust simulation states \rev{during rollout}, reducing the realism and adaptability of the simulation. These shortcomings highlight the need for more realistic and flexible autonomous driving simulation environments. This remains a key direction for current research \cite{9,10}. Existing traffic simulators lack mechanisms to dynamically align simulated vehicle behaviors with evolving real-world traffic patterns; this work addresses how to achieve such alignment through an LLM-augmented online calibration framework.

To address the above issues, \rev{we present an online calibration framework designed to improve the alignment between the simulated environment and real-world traffic data during rollout}. To counteract the degradation of environmental realism during simulation, we introduce a timed discrepancy detection mechanism and \rev{leverage an LLM} to inform and adjust vehicle decision-making. \rev{Experiments on selected HighD highway scenarios} demonstrate the effectiveness of our approach in preserving simulation validity, as shown in Section \ref{sec:REARLMethodValidation}. The contributions of this paper are as follows:

\begin{itemize}
    \item \rev{We formulate REARL as a closed-loop simulation calibration framework for autonomous driving simulation, which monitors distributional discrepancies during rollout and adjusts the simulation environment to improve alignment with real traffic statistics under the tested setting.}
    \item This approach \rev{integrates two main components}: the timed detection of traffic flow changes and representative scenarios-guided LLM decision-making for background vehicles. Traffic data undergoes clustering to extract representative scenarios based on cluster centers. Subsequently, these scenarios are utilized to support the decision-making and simulation processes in LLMs. The timed detection mechanism monitors vehicle speed distribution and mean spacing between pairs of vehicles. It triggers LLM intervention when thresholds are exceeded \rev{to support discrepancy-aware adjustment during simulation rollout}.
    \item Our experiments were conducted on the \rev{HighD dataset} \cite{22} and the highway-env simulation environment \cite{26}. Compared to the \rev{CRITICAL method}, our approach reduces the Hellinger distance for speed distributions to 0.3067 and the MAPE for mean spacing between pairs of vehicles to 0.8371.
\end{itemize}

The validity and effectiveness of the proposed REARL framework rely on four key assumptions. First, we assume that representative scenarios, \rev{derived from clustering real-world traffic data, capture essential patterns of dynamic traffic flow and provide realistic driving-context references}. Second, we posit that monitoring only two macroscopic indicators------vehicle speed distribution and mean spacing between background vehicle pairs------\rev{provides a tractable proxy for monitoring simulation fidelity}, as both are highly sensitive to traffic flow consistency. Third, we assume that a Large Language Model (LLM), when guided by scenario-informed prompts, can plausibly imitate driver behaviors from static observational snapshots \rev{under predefined efficiency and traffic-rule constraints}. Fourth, we assume that LLM-based interventions \rev{can reduce accumulated distributional deviations during simulation rollout under the tested setting, so that micro-level adjustments can contribute to macro-level realism}. Together, these assumptions underpin REARL's closed-loop, adaptive enhancement mechanism.

The framework continuously optimizes simulation states through a timed discrepancy detection and adjustment mechanism. This \rev{aims to improve simulation realism in the tested dynamic traffic environments}, providing a reliable platform for testing and validating autonomous driving systems. The structure of the following chapters is as follows: Section 2 presents related work on autonomous driving simulation methods and their limitations. Section 3 provides a detailed description of REARL method. Section 4 validates REARL from different perspectives. Section 5 concludes the paper and discusses future directions.

\section{Related works}
The rapid development of autonomous driving technology has raised higher demands for high-fidelity simulation environments. These are necessary to enhance the objectivity and credibility of the testing. Existing research mainly focuses on building realistic driving simulators. These can be categorized into three types: traditional simulation methods, data-driven methods, and the application of LLMs in autonomous driving. While these methods have advanced simulation technology, they still have significant limitations. These limitations are especially evident in statistical realism, dynamic adaptability, and modeling of complex scenarios.

\subsection{Traditional simulation methods}
Traditional simulation methods rely on predefined rules and heuristic models to simulate traffic behavior. Representative tools include SUMO \cite{23}, VISSIM \cite{24}, AIMSUN \cite{25}, and Highway-env \cite{26}. These methods typically use physics-based models, such as car-following models \cite{27,28} and lane-changing models \cite{29,30}. They have been studied for decades in the field of traffic engineering. However, these methods have limitations. Their parameterization and manually coded rules prevent them from capturing the dynamic characteristics and nonlinear interactions of real-world traffic. In high-density traffic or complex driving scenarios, they struggle to simulate time-dependent behavior. This results in significant differences between simulated and real driving environments \cite{31}. Furthermore, they do not provide the required precision for modeling safety-critical scenarios, such as collisions or dangerous situations. This limits the credibility and realism in autonomous driving testing.

\subsection{Data-driven simulation methods}
To overcome the limitations of traditional methods, data-driven approaches use large-scale real-world traffic data. They aim to improve simulation realism. Relevant techniques include direct sampling and clustering of traffic primitives \cite{32}. They also include probabilistic modeling using Bayesian networks \cite{33}. scenario enhancement with deep generative models is another technique \cite{34}. These methods have made progress in dynamic traffic representation. Neural networks \cite{35,36,37} are commonly used to model specific scenarios or vehicle behaviors. For example, Mo et al. \cite{38} proposed a data-driven framework that combines deep learning models with traffic flow simulation to improve prediction accuracy in complex environments. Similarly, Liu et al. \cite{39} introduced a learning-based stochastic driving model for autonomous vehicle testing, using a long short-term memory (LSTM) network to generate human-like, interactive vehicle behaviors. Furthermore, Liu et al. \cite{40} developed a data-driven simulation system (DDSS) that uses a Sim-Hybrid Retraining Constrained LSTM (SHRC-LSTM) model for traffic flow prediction, achieving higher precision than traditional methods like VISSIM in assessing efficiency, safety, and emissions. These networks model specific scenarios or vehicle behavior. However, data-driven methods face challenges. They include data bias, overfitting, and poor interpretability. Models that rely purely on data are prone to noise interference. This occurs when modeling long-tail safety-critical events, which can hinder the precise representation of rare yet high-impact traffic situations. It makes precise decision support difficult. Existing data-driven methods generate only short-term simulations. For example, a few seconds (e.g., D2Sim \cite{41}). This limits their applicability in complex interaction scenarios. It also limits their ability to meet the full journey training needs of autonomous driving.

\subsection{Application of LLMs in Enhancing Autonomous Driving Simulation Environments}
In recent years, large language models (LLMs) have been incorporated into autonomous driving due to their advanced reasoning and contextual understanding capabilities \cite{42,43}. \rev{Beyond autonomous driving, LLMs have also shown potential in wireless communication tasks, such as massive MIMO CSI feedback and large-small model collaboration for air-interface optimization \cite{llm_mimo_csi_feedback,llm_air_interface_collaboration}.} For instance, LinguaSim utilizes natural language instructions to generate realistic multi-vehicle test scenarios, thus enhancing the diversity of scenario libraries \cite{50}. Additionally, multi-agent LLM frameworks have been applied in design space exploration to enhance simulation efficiency \cite{51}. Comprehensive reviews emphasize the growing role of LLMs in autonomous driving systems, including reasoning, behavior modeling, and multi-task coordination, providing a solid theoretical and experimental foundation for integrating intelligent reasoning into simulations \cite{52,53,54}. \cite{21} proposed the use of LLMs to optimize simulation configurations by generating critical scenarios such as edge cases and boundary conditions. This process enhances the diversity of the training data. However, it only modifies the initial state of the simulation environment, ensuring that the starting conditions align with real traffic data. In the meantime, background vehicles in the simulation are controlled using a rule-based approach, with the parameters of the rule-based model adjusted based on statistical analysis of real datasets. However, as the simulation progresses, particularly in the later stages, the interactions between background and target vehicles may lead to deviations, resulting in uncontrolled scenarios, and the simulation may no longer remain consistent with real traffic data.

\section{Methodology}
\subsection{Overview}
In this study, we propose REARL, \rev{a closed-loop calibration framework designed to reduce distributional deviations in autonomous driving simulations by integrating real-world traffic data with LLM-guided intervention}. REARL aims to improve the realism and adaptability of autonomous driving environments, addressing the limitations of traditional simulation methods.

\begin{figure*}[htbp]
\centering
\includegraphics[width=0.85\linewidth]{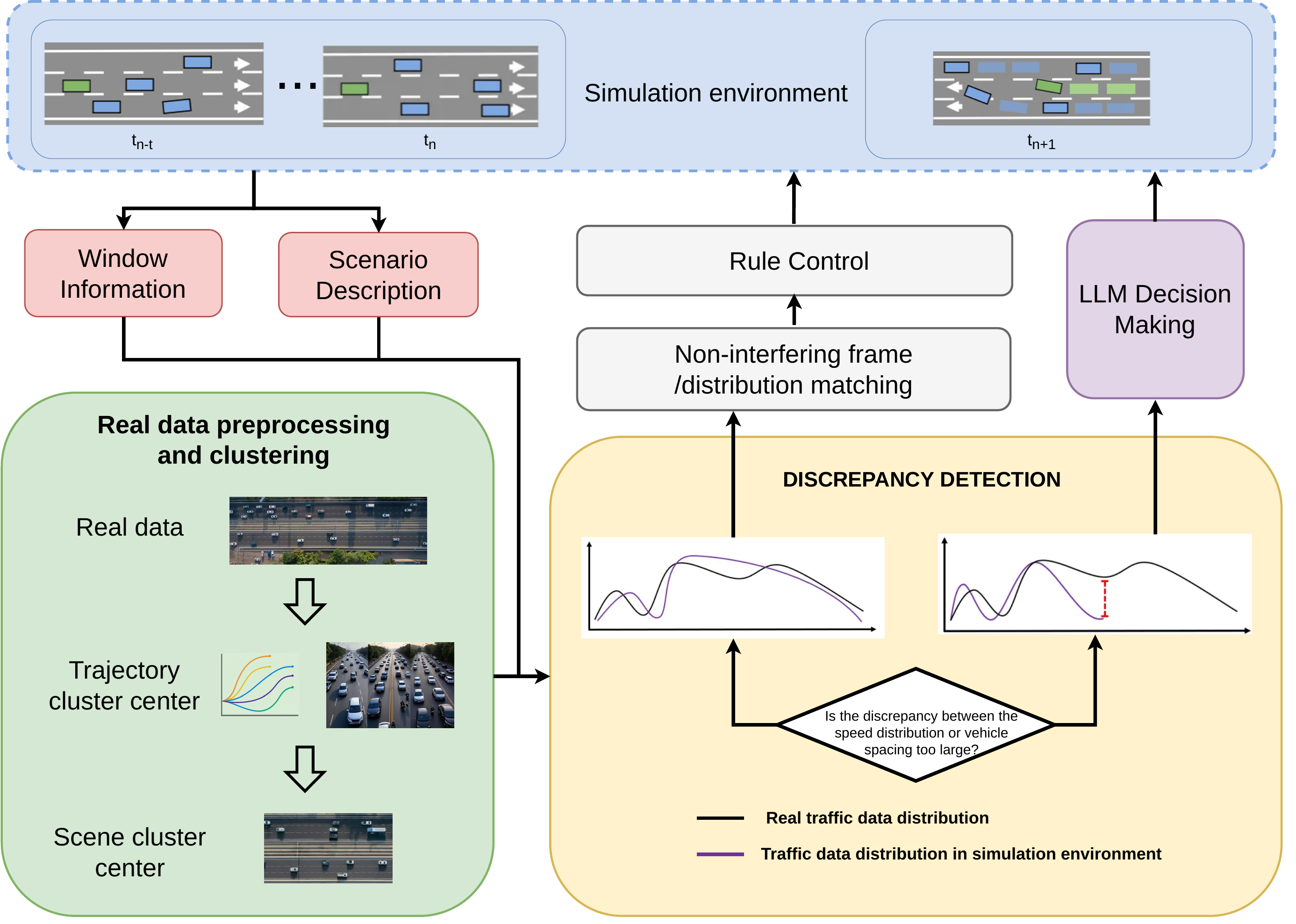}
\caption{The REARL framework consists of three main modules: data preprocessing, a timed detection framework and LLM decision-making. The timed detection framework continuously monitors the deviation between the simulation environment and real data. If a large deviation is detected, the LLM decision module is activated. \rev{It provides adjustment suggestions for vehicle control strategies to improve consistency between the simulated traffic state and the real-traffic reference.}
}
\label{fig2}
\end{figure*}

As \rev{shown} in Fig. \ref{fig2}, the framework consists of three interconnected modules: Real Data Preprocessing and Clustering, Speed Distribution and mean spacing between pairs of vehicles Discrepancy Detection, and representative scenarios-guided LLM Decision-Making for Vehicle Behavior Adjustment.

Real Data Preprocessing and Clustering: This module processes high-frequency vehicle trajectory data collected from real-world traffic scenarios. The data is clustered into representative traffic scenarios, capturing key patterns in vehicle movement and behavior. \rev{These clustered data sets serve as a foundation for the subsequent modules by providing representative real-traffic references for simulation adjustment.}

Speed Distribution and mean spacing between pairs of vehicles Discrepancy Detection: Building on the preprocessed and clustered data, this module continuously monitors the simulation's alignment with real traffic data. It focuses on detecting discrepancies in vehicle speed and spacing, which are critical indicators of realistic traffic flow. \rev{When significant deviations from real-world traffic statistics are identified, the system triggers interventions to adjust the simulation and improve its consistency with real traffic dynamics.}
Representative Scenarios-Guided LLM Decision-Making for Vehicle Behavior Adjustment: Upon detecting discrepancies, the LLM module intervenes by adjusting the vehicle behavior within the simulation. It uses insights derived from the clustered real-world data \rev{to generate adjustment decisions during simulation rollout, with the aim of improving alignment between simulated and observed traffic patterns}. \rev{This dynamic adjustment process allows REARL to improve simulation realism during rollout by adapting to observed traffic-flow discrepancies, while its effectiveness remains bounded by the evaluated scenarios and metrics.}

More details about the Real Data Preprocessing and Clustering, Speed Distribution and mean spacing between pairs of vehicles Discrepancy Detection, and Representative Scenarios-Guided LLM Decision-Making for Vehicle Behavior Adjustment modules will be elaborated in the following sections.

\subsection{Real data preprocessing and clustering}
\label{sec:dataPreprocessing}
In this study, we perform cluster analysis on vehicle trajectories based on the \rev{HighD dataset}. The HIGH D dataset contains high-frequency vehicle trajectory data, sampled at 25 frames per second \cite{22}. To ensure the data is suitable for simulation environment analysis, we align the real data time scale with the simulation environment and normalize the speed values using min-max normalization to the range [5, 32]. Each simulation cycle has a duration of $T_s = 60$ seconds, with decisions made once per second. Therefore, the real traffic data is trimmed to create 60-second traffic scenarios $s_i$ for the simulation environment, retaining only the snapshot data for each second. A total of 741 traffic scenarios were obtained. The data trimming process meets the following constraints: For any time scale $t \in [0, T_s]$, the number of vehicles in the scenario $N_v(t) \geq 10$. This excludes noisy scenarios. Therefore, the trimmed dataset $S' \subseteq D$ satisfies Eq. \ref{eq:1}. Since the number of vehicles recorded per frame in the real data varies, we select the 10 vehicles that have been recorded for the longest duration in each frame as the representative snapshot for that second.

\begin{figure*}[htbp]
\centering
\includegraphics[width=1\textwidth]{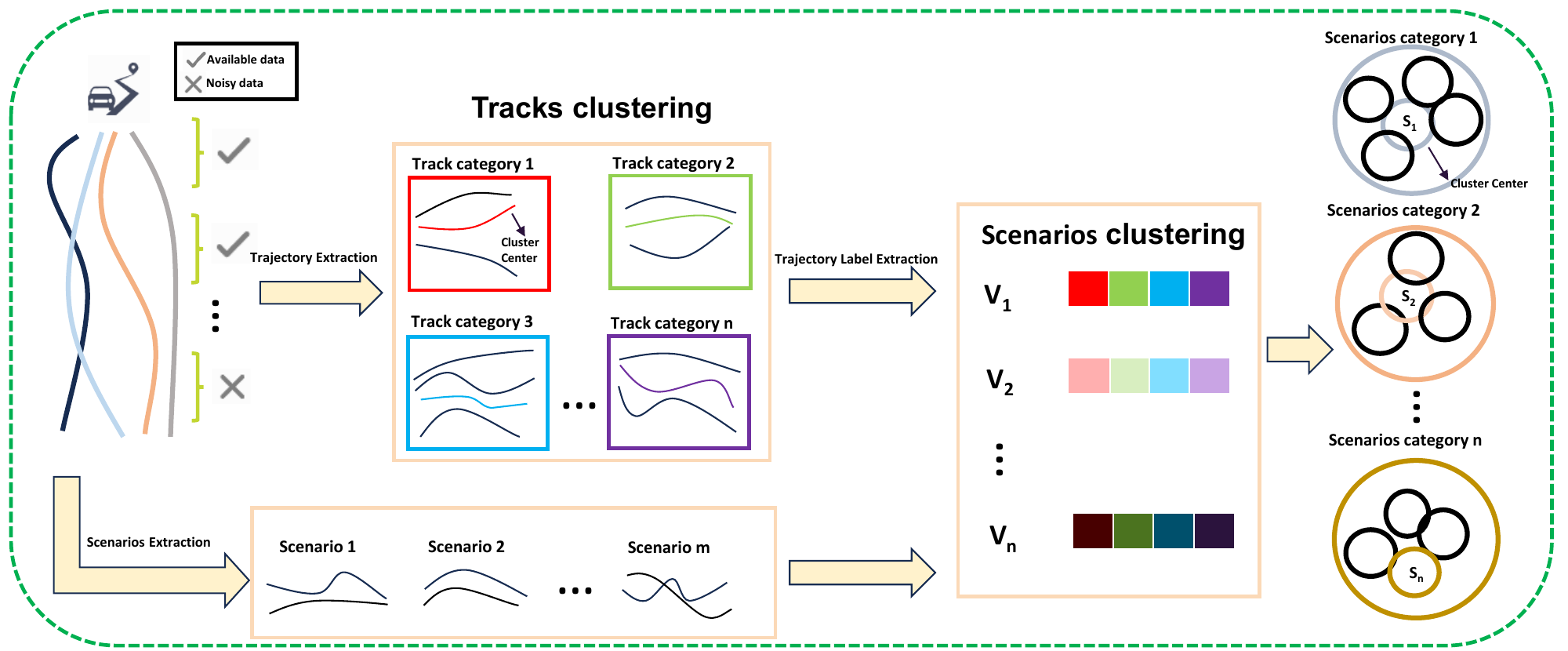}
\caption{
Clustering method. Track category n represents each cluster after clustering. For example, in track category 1, the red track represents the cluster center. $V_n$ is a characterization vector for each scenario, indicating the proportion of each track type within the scenario. Scenario category n represents the result of scenario clustering. For example, in Scenario category 1, the gray scenario $S_1$ represents the cluster center.
}
\label{fig3}
\end{figure*}

\begin{equation}
S' = \{ s_i \in D \mid \forall t \in [0, T_s], N_v(t) \geq 10 \}
\label{eq:1}
\end{equation}

Fig. \ref{fig3} illustrates the data clustering process. We first extract the trajectories $D$ from the \rev{HighD dataset}. To ensure the reliability of the clustering results, we filter out trajectories with length $L(P_i) < 100$ time steps. As shown in Eq. \ref{eq:2}, the final trajectory set is denoted as $P={P_1, P_2, ..., P_n}$, where $P_i$ represents the i-th vehicle trajectory and $n$ is the total number of trajectories. \rev{Trajectory clustering} is then performed using the center-based kicluster algorithm \cite{45} to obtain trajectory labels $Q$. The scenario representation $V$ is calculated, where the representation of each scenario is a vector formed by the proportions of each type of trajectory in that scenario. Subsequently, k-means clustering is applied to obtain the scenario centers $M$. This figure visually presents the data flow and the core logic of the method.

\begin{equation}
P = \{ P_i \in P' \mid L(P_i) \geq 100 \}
\label{eq:2}
\end{equation}
Here, $L(P_i)$ represents the number of time steps in trajectory $P_i$. Each trajectory $P_i$ has a feature vector $F(P_i(t)) = (v_i(t), x_i(t), d_i(t))$ at each time step $t$. These represent speed, position, and spacing, respectively. To quantify the similarity between trajectories, we use the approximating Fr\'echet distance as a metric \cite{46}.

Based on kicluster algorithm trajectory clustering results $Q$, we further perform scenario clustering, grouping the scenarios into 15 categories. For each scenario $s_i$, we use vehicle ID labels to calculate vehicle type ratios. This generates a feature vector $v_k \in \mathbb{R}^{3}$ with a length of 3. We apply $k$-means clustering to the scenario feature set $V = \{v_1, v_2, ..., v_k\}$. The goal is to minimize the within-cluster squared distance, as shown in Eq. \ref{eq:6}.
\begin{equation}
J = \sum_{k=1}^K \sum_{j=1}^{15} r_{kj} \| v_k - v_{u_j} \|^2
\label{eq:6}
\end{equation}
Here, $r_{kj} = 1$ if $v_k$ belongs to cluster $j$, otherwise $r_{kj} = 0$. $v_{u_j}$ is the central scenario of cluster $j$. The resulting scenario cluster centers $M = \{s_{u_1}, s_{u_2}, ..., s_{u_{15}}\}$ represent representative traffic scenarios.

\subsection{A speed distribution and mean spacing between pairs of vehicles discrepancy detection framework}
\begin{figure}[!t]
\centering
\includegraphics[width=1\linewidth]{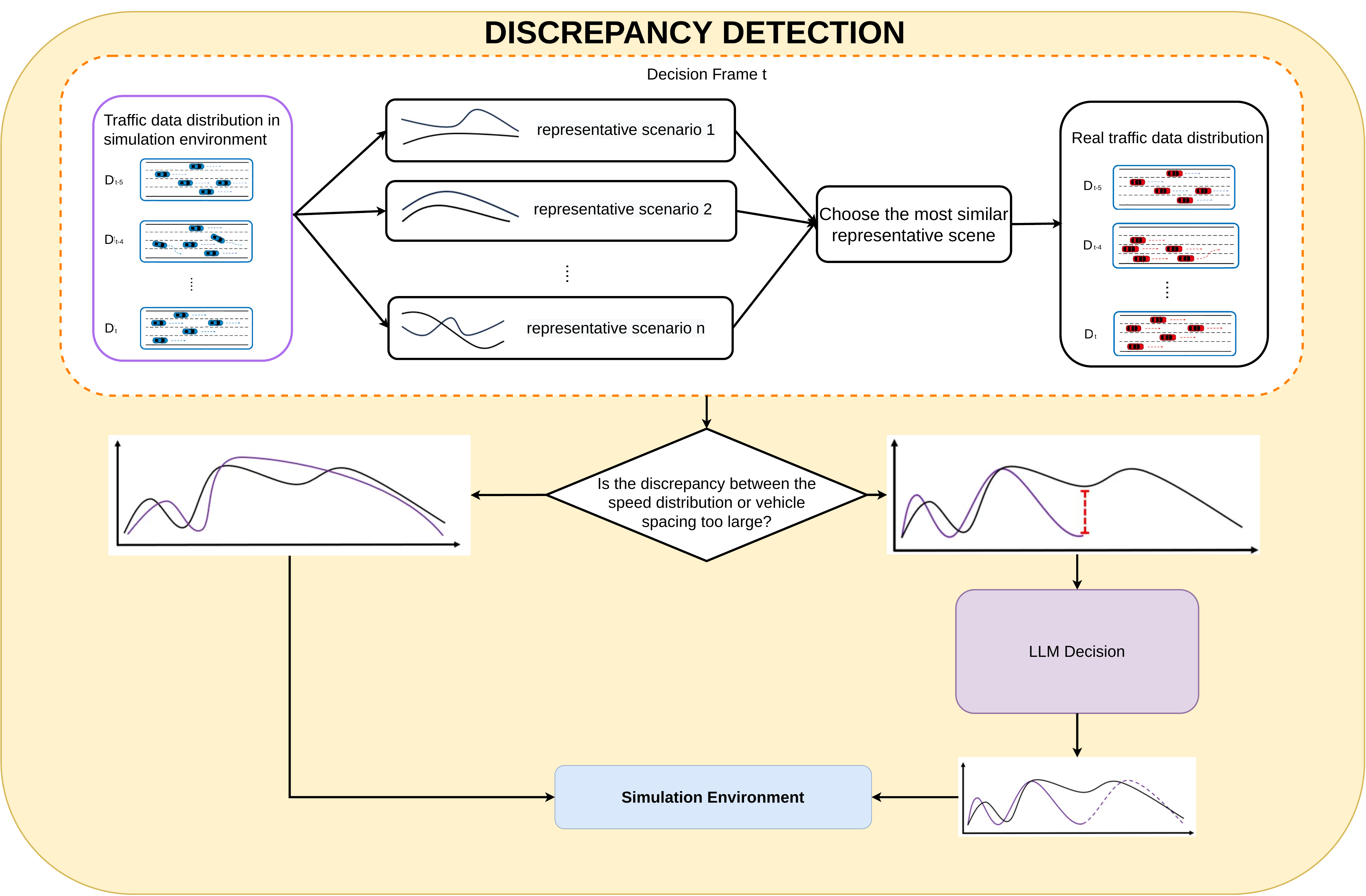}
\caption{Speed distribution and mean spacing discrepancy detection framework. At each detection point, we select a reference scenario. It is the representative scenario with the smallest average frame-wise MAPE of mean spacing over a prefix window. This MAPE is computed between the mean spacing of all background vehicle pairs in the simulation and that in each candidate scenario. Using this reference, we evaluate two discrepancies. One is the Hellinger distance of speed distributions over the past 5 seconds. The other is the MAPE of current-frame mean spacing. If either exceeds a threshold, LLM intervention is triggered. The LLM then adjusts vehicle control strategies in the simulation.}
\label{fig4}
\end{figure}

\rev{To support more consistent simulated interactions when LLM decision-making is involved, we introduce a discrepancy detection mechanism based on speed distribution and mean spacing between vehicle pairs.} This mechanism determines whether LLM decision-making intervention is required. First, we identify the most similar representative scenario by averaging the frame-wise MAPE of mean spacing over a prefix window, where the MAPE at each frame compares the mean spacing of all background vehicle pairs in the simulation with that in each candidate scenario. The scenario with the smallest MAPE is selected as the reference. Using this reference, we compute the Hellinger distance between the speed distributions over the past 5 seconds in the simulation and real traffic data. We also compute the MAPE of mean spacing in the current frame. When significant deviations are detected between the simulation's statistics and real data, the timed detection mechanism triggers LLM intervention. \rev{This provides an opportunity to adjust the vehicle control strategies in the simulation environment.} The process is shown in Fig. \ref{fig4}.

Specifically, the speed distribution and mean spacing between pairs of vehicles discrepancy detection mechanism makes judgments based on the following formulas. The Hellinger distance is used to measure the similarity between the vehicle speed distribution in the simulation environment and the real data speed distribution. The collected vehicle speed data are binned and statistically analyzed according to the intervals [ 5.0, 8.375, 11.75, 15.125, 18.5, 21.875, 25.25, 28.625, 32.0 ]. Compared to KL divergence, the Hellinger distance has better symmetry and stability in its calculation, making it more precise in reflecting the similarity between real data and simulation data \cite{8}. Its formula is shown in Eq. \ref{eq:7}.
\begin{equation}
H(P, Q) = \frac{1}{\sqrt{2}} \left( \sum_{i=1}^n \left( \sqrt{P_i} - \sqrt{Q_i} \right)^2 \right)^{1/2}
\label{eq:7}
\end{equation}

Here, $P_i$ and $Q_i$ represent the speed distributions in the simulation environment and real data, respectively. The MAPE is used to assess how closely the mean spacing between pairs of vehicles in the simulation environment matches real traffic data. While RMSE gives more weight to larger errors and MAE treats all errors equally, MAPE provides a clearer and more balanced indication of model performance. \cite{49} Unlike RMSE and MAE, MAPE is scale-independent and calculates the average absolute percentage difference, making it particularly suitable for comparing simulations to real traffic data. This is crucial in our case, as each scenario and frame in both simulation and real traffic data may have different scales. By using MAPE, we aim to fairly quantify the relative realism of the simulation, measuring how closely it approximates real-world traffic data across varying scenarios and frames. The formula is shown in Eq. \ref{eq:8}.

\begin{equation}
\text{MAPE} = \frac{1}{n} \sum_{i=1}^n \left| \frac{y_i - \hat{y}_i}{y_i} \right|
\label{eq:8}
\end{equation}

Here, $\hat{y}_i$ represents the real mean spacing between pairs of vehicles, and $y_i$ is the mean spacing between pairs of vehicles in the simulation environment. When the Hellinger distance or MAPE values exceed the set threshold, the timed detection mechanism notifies the LLM to make adjustments. \rev{This intervention is intended to improve the alignment between simulated vehicle behavior, traffic-flow statistics, and the corresponding real-data reference.}

\subsection{Representative scenarios-guided LLM decision-making for background vehicles}
\begin{figure*}[htbp]
\centering
\includegraphics[width=\textwidth]{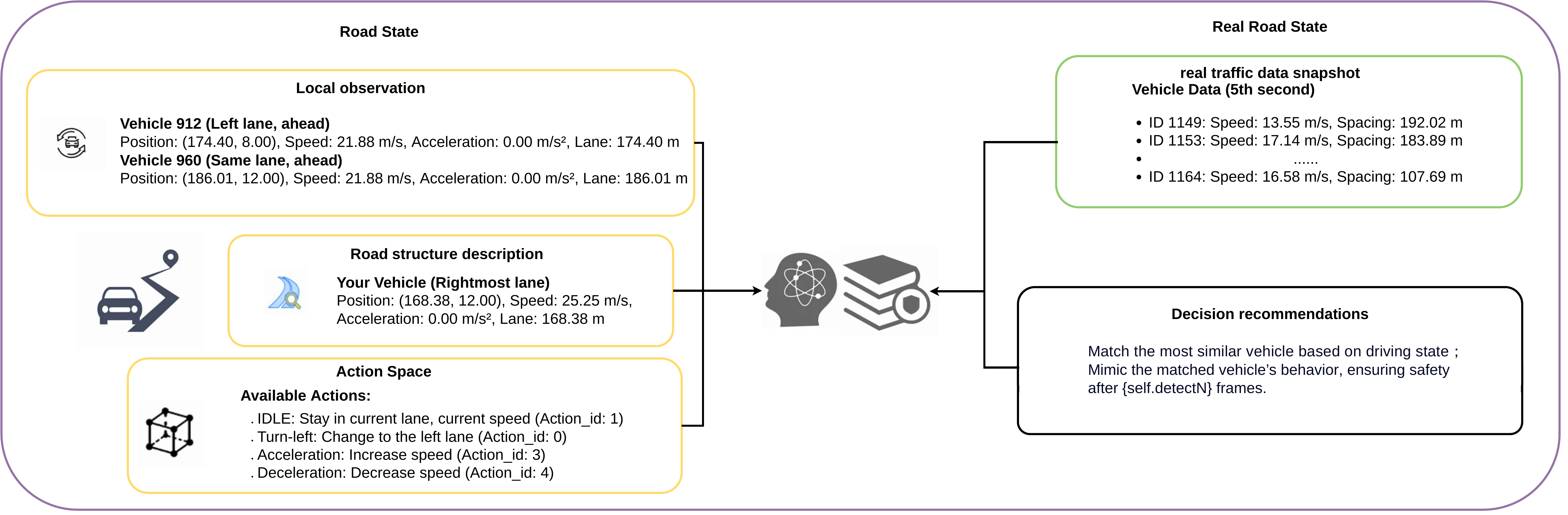}
\caption{
LLM Decision Design. It consists of three core components. First, prefix comparison is used to compare the representative scenarios with the current simulation data, selecting the representative scenario with the smallest difference. The snapshot of the current moment from the representative scenario is selected as the real-world traffic data snapshot, providing real-world traffic conditions as a reference for LLM decision-making. Second, the road description module provides a detailed description of the current road network structure, the vehicle's location on the network, and its driving status. Finally, the decision-making module focuses on the design of prompt to optimize the decision-making and reasoning process.}
\label{fig5}
\end{figure*}

To enhance the simulation environment, we propose a method that leverages LLMs and representative scenarios for decision-making of background vehicles. \rev{This method is designed to guide the simulation toward the statistical characteristics of real datasets, thereby improving realism under the evaluated setting.} This method is inspired by the DILU framework \cite{8}. It uses LLM to control vehicle decision-making, simulating more realistic traffic scenarios. Fig. \ref{fig5} shows the specific decision prompt design.

The real traffic data snapshots originate from a set of representative scenarios, which represent the most typical traffic patterns. These scenarios are defined by the cluster centers $M = \{s_{u_1}, s_{u_2}, ..., s_{u_{15}}\}$, as derived from the clustering process described in Section \ref{sec:dataPreprocessing}. For all representative scenarios, at simulation moment $t$, we compare the simulation data from time 0 to $t$ with the data of all representative scenarios from time 0 to $t$ using prefix comparison, calculating the differences in vehicle speed distribution and the mean spacing between pairs of vehicles. The representative scenario with the smallest difference is selected. We then extract the speed and spacing of the top 10 vehicles with the longest driving time from the frame corresponding to time moment $t$ of the representative scenario. These serve as feature representations of road conditions. For any scenario center $u_k$, the feature vector at the decision frame is defined as shown in Eq. \ref{eq:9}.
\begin{equation}
F_{u_k}(t) = (v_1(t), v_2(t), \dots, v_{10}(t), d_1(t), d_2(t), \dots, d_{10}(t))
\label{eq:9}
\end{equation}
Here, $v_i(t)$ represents the speed of the $i$-th vehicle, and $d_i(t)$ represents the average spacing with the other 9 vehicles. Through prompt engineering design, this information serves as a reference for decision-making, allowing the LLM to learn the dynamic characteristics of real traffic flow.

The road description module mainly includes descriptions of the current road network, the vehicle's position within the road network, and the vehicle's current driving state. In real traffic systems, drivers make decisions based on the behavior of surrounding vehicles. Therefore, the local observation information module provides the driving status of other vehicles nearby. Based on the local observation information provided by the highway-env simulation platform, it is converted into textual form. The action description module calculates the vehicle's action space based on its current position and surrounding road conditions. This reduces interference noise and \rev{reduces the likelihood of ineffective or inadmissible decisions}. \rev{In implementation, the LLM output is parsed as a discrete action identifier within the predefined action space; outputs that cannot be parsed or fall outside the admissible set are treated as invalid and are not directly applied to the simulator.}

The decision suggestion module primarily involves the design of prompts to optimize the decision-making process. At each simulation time step $t$, the LLM generates decisions based on the snapshot $F_{u_k}(t)$. \rev{When an invalid or inadmissible action is detected, the framework falls back to the existing rule-based control strategy for that vehicle at the current step.}
\rev{For lane-changing actions, the admissibility check further considers whether the target lane is available in the current road context before the action is accepted.}
\rev{This mechanism is intended as an action-level validity filter rather than a formal safety verification guarantee.} This module focuses solely on optimizing the simulation to closely approximate real traffic data. \rev{The goal is to adjust the simulation environment---specifically, the speed distribution and mean spacing between vehicle pairs---to reduce discrepancies relative to real-world traffic patterns.} Moreover, the simulation not only aims to match real-world traffic data but also incorporates safety considerations for background vehicles, \rev{discouraging collision-prone maneuvers} and that \rev{vehicle behaviors are constrained by the admissible action space}.

The method derives representative scenarios from real traffic data via preprocessing and clustering, identifies the one most similar to current conditions through prefix comparison, and produces time aligned snapshots $F_{u_k}(t)$. Then, it employs prompt-designed LLM decision-making guided by snapshot $F_{u_k}(t)$ to jointly optimize traffic realism and \rev{promote safer behavior of background vehicles}. Since decisions directly affect the simulation environment, the traffic flow simulation's speed and spacing distributions $P_{sim}$ are gradually optimized. Ultimately, they approximate the real data distribution $P_{real}$. \rev{This process forms a detection-adjustment feedback loop that supports the realism and dynamic consistency of the simulation environment.}

\section{Experiments}
\subsection{Experimental settings}

Our experiment uses a well-established highway-env as the simulation platform, widely used in autonomous driving and tactical decision-making research \cite{26}. This environment offers a realistic multi-vehicle interaction setup and allows direct modification of the underlying code, providing a flexible experimental framework. The simulation is conducted in a four-lane highway environment, \rev{with 10 background vehicles, a vehicle density of 2.0, and a simulation horizon of 60 steps}.

The validation is performed using four evaluation metrics: the Hellinger distance, the MAPE, the average time headway (THW), and the lane change rate. The Hellinger distance and the MAPE capture macroscopic distributional similarity between the simulation and real traffic data. Specifically, the Hellinger distance measures the difference between the speed distribution of real traffic data and the simulation environment's speed at each decision frame over a 5-second window. The final result is the average Hellinger distance across all decision frames. The MAPE calculates the mean distance between all vehicles and every other vehicle on the road at each decision frame. It then compares the average mean spacing between pairs of vehicles of real traffic data with that of the simulation environment, with the final result being the average MAPE across all decision frames. To further evaluate microscopic traffic-flow fidelity, we introduce THW and lane change rate (LCR). THW is a commonly used longitudinal \rev{car-following indicator} in traffic engineering, defined as the ratio of the spatial gap between a leader--follower vehicle pair in the same lane to the follower's speed, measured in seconds. A larger THW indicates \rev{more conservative car-following behavior}. For each episode, we compute the mean THW across all decision frames and all valid same-lane leader--follower pairs as the longitudinal \rev{behavioral metric}. LCR quantifies the frequency of lateral maneuvering behavior. A successful lane change is defined as a change in a vehicle's lane ID between two adjacent decision frames without a collision in the target frame, thereby excluding lane-change attempts that fail due to a crash. The episode-level LCR is computed as the total number of successful lane changes divided by the product of the number of decision frames and the number of vehicles, normalized to the range $[0, 1]$. A higher LCR indicates more frequent lane-changing and \rev{a stronger tendency for lateral maneuvering}. Together, these four metrics provide a \rev{broader but still bounded} assessment of simulation realism from both macroscopic distributional alignment and microscopic behavioral fidelity perspectives.

\subsection{The validation of the REARL method}
\label{sec:REARLMethodValidation}
In this section, we evaluate the proposed method against three baselines and three variants of REARL on \rev{50 randomly selected HighD highway scenarios}. The compared methods include a rule-based baseline (Base), a reinforcement learning baseline (PPO), an LLM-enhanced simulation baseline (CRITICAL) \cite{21}, and three REARL variants with different backbone models, namely Qwen3 32B, DeepSeek-R1 32B, and GLM-4.7-Flash 30B. \rev{To ensure a controlled comparison under the same highway setting}, all methods are evaluated under the same \rev{controlled highway simulation setting}, a four-lane highway with 10 controlled vehicles, a vehicle density of 2.0, and a simulation horizon of 60 steps. Each method is evaluated over five independent runs. \rev{For all LLM-based variants, we use deterministic decoding with the sampling temperature set to 0, so that the same prompt and simulation state produce reproducible action outputs.} \rev{The prompt template follows the structure shown in Fig. \ref{fig5}, and the model is instructed to return only the selected discrete action identifier from the predefined action space.} \rev{The same parsing, admissibility-checking, and fallback rules described above are applied consistently during evaluation.} In each run, we compute the Hellinger distance between speed distributions, the mean absolute percentage error (MAPE) of the average spacing between pairs of background vehicles, the average time headway (THW), and the lane change rate on \rev{the same 50 selected HighD highway scenarios}. The final performance is obtained by averaging the results across all runs and scenarios. \rev{The 50 evaluation scenarios are sampled from the held-out scenario subset rather than from the subset used for threshold estimation.}


\begin{table*}[htbp]
\centering
\caption{Comparison of effectiveness with other methods}
\label{table:1}
\resizebox{\textwidth}{!}{
\begin{tabular}{l c c c c}
\toprule
\textbf{Method}
& \textbf{Speed Distribution}
& \textbf{Mean Spacing}
& \textbf{THW}
& \textbf{Lane Change Rate} \\
& \textbf{(Hellinger Distance)}
& \textbf{(MAPE)}
&  &  \\
\midrule
Base
& 0.3460
& 0.8978
& 7.6679
& 0.0081 \\

PPO
& 0.3258
& 1.3940
& 10.6617
& 0.0138 \\

CITICAL
& 0.3277
& 2.0523
& 7.2094
& 0.5630 \\

REARL (DeepSeek-R1 32B)
& \textbf{0.2562}
& \underline{0.8600}
& \underline{16.3352}
& 0.0616 \\

REARL (GLM-4.7-Flash 30B)
& 0.3131
& 1.2099
& 14.2384
& 0.0617 \\

REARL
& \underline{0.3067}
& \textbf{0.8371}
& \textbf{22.8575}
& 0.0708 \\
\bottomrule
\end{tabular}
}
\end{table*}

\begin{itemize}[
  label=$\scalebox{0.75}{$\bullet$}$,
  itemsep=0pt,
  parsep=0pt,
  topsep=0pt
]
    \item \textbf{Base}: This baseline uses the rule-based models provided by highway-env under the same simulation setting. Specifically, IDM (Intelligent Driver Model) is used to model longitudinal driving behavior, whereas MOBIL (Minimizing Overall Braking Induced by Lane Changes) is used to model lateral lane-changing decisions.

    \item \textbf{PPO}: This baseline employs Proximal Policy Optimization (PPO) \cite{55}, a widely used deep reinforcement learning algorithm, to train a policy that controls the longitudinal and lateral behavior of background vehicles. PPO is trained under the same simulation setting as the other methods. Including PPO as a learning-based baseline allows us to assess whether a general-purpose RL policy, optimized purely through environment interaction, can achieve comparable simulation realism without \rev{direct real-data guidance} or LLM-based reasoning.

    \item \textbf{CRITICAL} \cite{21}: This method leverages LLMs to optimize the initial simulation configuration, thereby improving the realism of the generated traffic environment. To ensure comparability, we adapt the original setting of CRITICAL to a 60-step simulation on a four-lane highway with 10 controlled vehicles and a vehicle density of 2.0.

    \item \textbf{REARL(DeepSeek-R1 32B)}: This variant uses DeepSeek-R1 32B as the backbone model within REARL. DeepSeek-R1 32B is a distilled reasoning model derived from DeepSeek-R1 based on the Qwen-32B model and is specifically optimized for reasoning-intensive tasks. Its inclusion allows evaluation of whether a reasoning-oriented backbone can further improve the proposed simulation enhancement framework.

    \item \textbf{REARL(GLM-4.7-Flash 30B)}: This variant uses GLM-4.7-Flash as the backbone model within REARL. GLM-4.7-Flash is a lightweight model in the GLM-4.7 series and emphasizes enhanced coding capability, more stable multi-step reasoning, and efficient execution in complex agent-style tasks. In addition, it supports a long context window, which makes it well suited to handling structured scenario descriptions and interactive simulation generation.

     \item \textbf{REARL(Qwen3 32B)}: This variant uses Qwen3 32B as the backbone model within REARL. Qwen3 represents the latest generation of the Qwen family and is designed to provide strong reasoning ability, instruction-following ability, agent capability, and multilingual support across both dense and MoE variants. The 32B dense version provides a strong and stable general-purpose LLM backbone, making it a suitable default choice for evaluating the effectiveness of REARL.
\end{itemize}

\begin{figure*}[!t]
\centering
\includegraphics[width=1\textwidth]{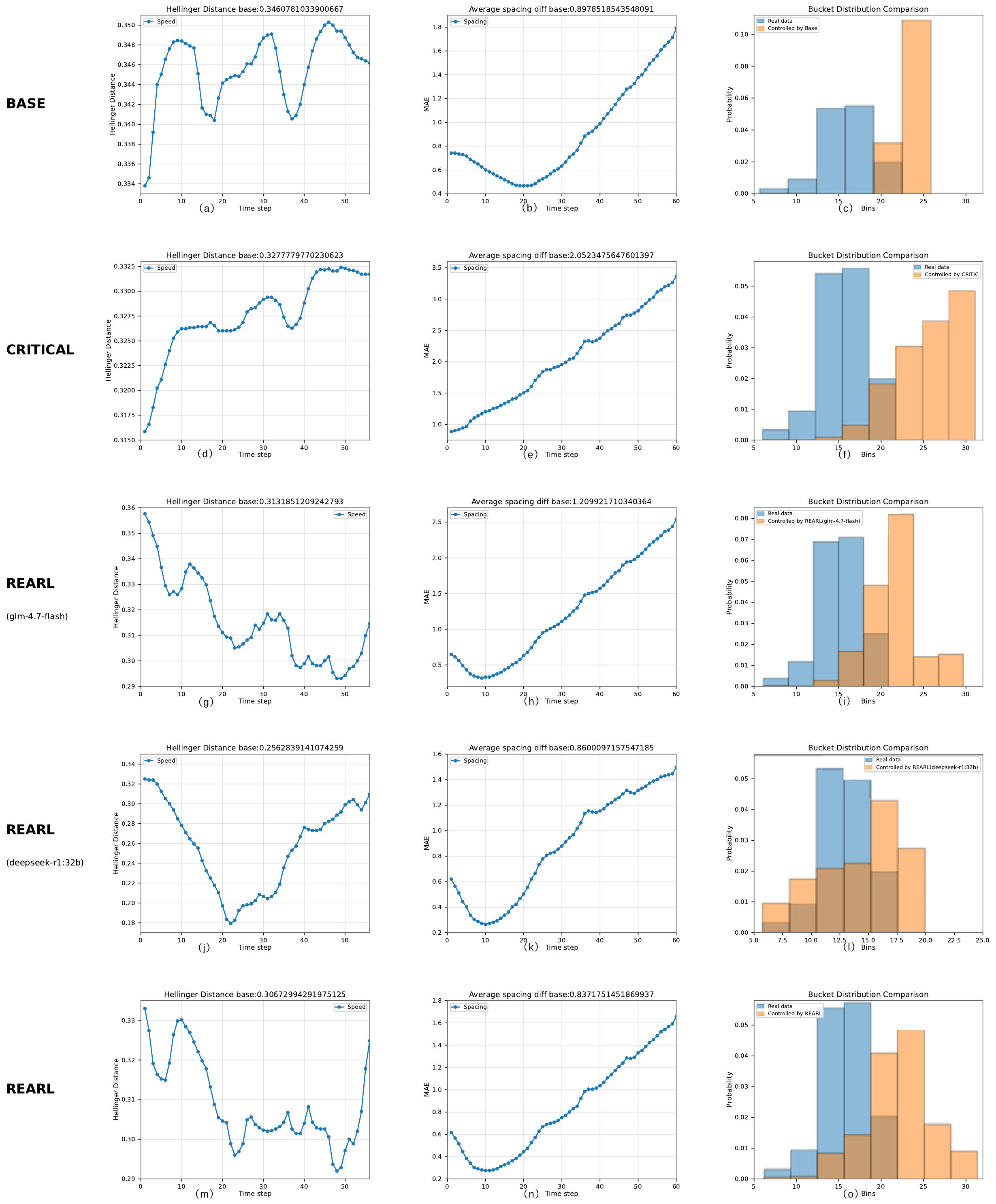}
\caption{Comparison of the simulation results generated by Base, CRITICAL, and REARL with different LLM backbones. From top to bottom, the rows correspond to Base, CRITICAL, REARL (glm-4-flash 30b), REARL (deepseek-r1 32b), and REARL (Qwen 32b), respectively. From left to right, the columns present the Hellinger distance of the speed distribution, the evolution of the mean spacing between vehicle pairs over time, and the histogram comparison of the overall speed distribution during the simulation.}
\label{fig6}
\end{figure*}

The quantitative results are summarized in Table \ref{table:1}. In terms of macroscopic distributional metrics, all REARL variants achieve lower Hellinger distances than the three baselines, with REARL(deepseek-r1 32b) obtaining the smallest value of 0.2562 and the default REARL achieving 0.3067. Among the baselines, PPO achieves a Hellinger distance of 0.3258, which improves over Base (0.3460) and is comparable to CRITICAL (0.3277), indicating that RL-based policy learning can partially reduce speed distribution discrepancy. However, for mean spacing MAPE, PPO yields a value of 1.3940, which is considerably worse than Base (0.8978) and only slightly better than CRITICAL (2.0523). \rev{This indicates that PPO does not achieve balanced alignment across the two macroscopic metrics under the current experimental setting. CRITICAL shows a similar trade-off: it reduces the Hellinger distance from 0.3460 for Base to 0.3277, but increases the spacing MAPE from 0.8978 to 2.0523. The default REARL achieves the lowest MAPE of 0.8371, and REARL(DeepSeek-R1 32B) achieves 0.8600, whereas REARL(GLM-4.7-Flash 30B) yields 1.2099, which is better than PPO and CRITICAL but worse than Base. Therefore, the results should not be interpreted as every REARL variant outperforming all baselines on every metric. Instead, they reveal a backbone- and metric-dependent trade-off, with the default REARL providing the most balanced performance across the two macroscopic metrics in this experiment.}

In terms of microscopic behavioral metrics, the advantages of REARL become more pronounced. For THW, a higher value generally indicates more conservative and safer car-following behavior with larger inter-vehicle time gaps. The default REARL achieves a THW of 22.8575, which is substantially higher than those of Base (7.6679), PPO (10.6617), and CRITICAL (7.2094). PPO's THW of 10.6617 represents a moderate improvement over Base and CRITICAL, suggesting that the RL policy learns to adopt slightly larger following gaps through environment interaction, but still falls considerably short of \rev{the larger inter-vehicle time gaps observed for REARL}. REARL(deepseek-r1 32b) and REARL(glm-4.7-flash 30b) also produce considerably higher THW values of 16.3352 and 14.2384, respectively, further confirming \rev{that the larger THW values are consistently observed across different REARL backbones under the current experimental setting.} For lane change rate, Base relies on the MOBIL rule-based model and yields an extremely low rate of 0.0081, indicating overly passive lateral behavior where vehicles rarely change lanes even when beneficial. PPO achieves a similarly low lane change rate of 0.0138, indicating that RL training under a standard reward formulation does not encourage realistic lane-changing behavior. In contrast, CRITICAL produces an excessively high lane change rate of 0.5630, suggesting unstable and erratic lateral decision-making that is inconsistent with realistic highway driving patterns. The three REARL variants achieve moderate lane change rates of 0.0616, 0.0617, and 0.0708, respectively, which reflect a more balanced and naturalistic level of lane-changing activity. \rev{These results indicate that REARL improves macroscopic distributional similarity and provides more conservative car-following and moderate lane-changing observations under the evaluated setting, but they should be interpreted as bounded behavioral evidence rather than comprehensive validation of simulation realism or driving safety.}

The temporal evolution of the speed-distribution discrepancy is shown in Fig. \ref{fig6}(a), Fig. \ref{fig6}(d), Fig. \ref{fig6}(g), Fig. \ref{fig6}(j), and Fig. \ref{fig6}(m). As shown in Fig. \ref{fig6}(a) and Fig. \ref{fig6}(d), Base and CRITICAL maintain relatively large Hellinger distances throughout the simulation, which indicates limited ability to stay aligned with real traffic dynamics over time. CRITICAL begins with a slightly smaller discrepancy than Base, suggesting that optimizing the initial simulation configuration provides a limited benefit at the early stage. However, this advantage gradually disappears as the rollout continues. By contrast, the three REARL variants in Fig. \ref{fig6}(g), Fig. \ref{fig6}(j), and Fig. \ref{fig6}(m) maintain lower speed-distribution discrepancies over much of the simulation horizon. Among them, REARL(deepseek-r1 32b) in Fig. \ref{fig6}(j) shows the largest reduction in Hellinger distance, while the default REARL in Fig. \ref{fig6}(m) also keeps the discrepancy consistently low and shows more balanced behavior over the full rollout. The REARL(glm-4.7-flash 30b) in Fig. \ref{fig6}(g) also improves clearly over Base and CRITICAL, although its gain is smaller than that of the other two REARL variants.

The histogram comparisons in Fig. \ref{fig6}(c), Fig. \ref{fig6}(f), Fig. \ref{fig6}(i), Fig. \ref{fig6}(l), and Fig. \ref{fig6}(o) further support these observations. As shown in Fig. \ref{fig6}(c) and Fig. \ref{fig6}(f), the speed distributions generated by Base and CRITICAL are mainly concentrated in the higher-speed region and show limited overlap with the real-data distribution. In contrast, the three REARL variants in Fig. \ref{fig6}(i), Fig. \ref{fig6}(l), and Fig. \ref{fig6}(o) produce distributions that are visibly closer to the real distribution, with much larger overlap. \rev{In particular, REARL(DeepSeek-R1 32B) in Fig. \ref{fig6}(l) shows the closest match in terms of speed-distribution similarity, while the default REARL in Fig. \ref{fig6}(o) also shows a favorable balance across the two macroscopic metrics. These results suggest that the proposed discrepancy-aware regulation strategy can reduce speed-distribution discrepancies relative to Base and CRITICAL under the tested backbone settings.}

A similar trend appears in the mean spacing between vehicle pairs, as shown in Fig. \ref{fig6}(b), Fig. \ref{fig6}(e), Fig. \ref{fig6}(h), Fig. \ref{fig6}(k), and Fig. \ref{fig6}(n). Although the spacing error increases over time for all methods because of accumulated rollout deviation, the increase is much more severe for CRITICAL in Fig. \ref{fig6}(e), which is consistent with its high MAPE of 2.0523 in Table \ref{table:1}. Base in Fig. \ref{fig6}(b) performs better than CRITICAL, but still shows a clear upward trend as the simulation proceeds. \rev{The three REARL variants exhibit different spacing-error trajectories in Fig. \ref{fig6}(h), Fig. \ref{fig6}(k), and Fig. \ref{fig6}(n), indicating that their spacing performance depends on the backbone model.} Notably, the default REARL in Fig. \ref{fig6}(n) achieves the lowest MAPE among all compared methods, indicating the best overall balance between speed-distribution alignment and spacing consistency. Although REARL(deepseek-r1 32b) in Fig. \ref{fig6}(k) achieves comparable spacing accuracy and the best speed-distribution similarity, the default REARL remains the most favorable setting when both metrics are considered together. \rev{REARL(GLM-4.7-Flash 30B) achieves a spacing MAPE of 1.2099, which is lower than PPO (1.3940) and CRITICAL (2.0523), but higher than Base (0.8978). Therefore, the improvement in spacing fidelity is not uniform across LLM backbones and should be interpreted together with the speed-distribution results.}

\rev{Overall, the results in Table \ref{table:1} and Fig. \ref{fig6}(a)--(o) show that REARL reduces the speed-distribution discrepancy across all three backbone models, whereas its spacing performance varies across backbones. Unlike CRITICAL, which adjusts only the initial simulation configuration, and PPO, which learns a policy without direct real-data guidance, REARL performs discrepancy-aware online adjustment during rollout. THW and lane change rate provide complementary descriptions of microscopic behavior. Under the current setting, the REARL variants produce larger THW values and intermediate lane change rates relative to the compared baselines; however, these observations should not be interpreted as direct validation of driving safety or universal behavioral realism. The results therefore demonstrate method- and metric-dependent trade-offs rather than uniform superiority across all four metrics. Among the three REARL variants, the default REARL based on Qwen3 32B provides the most balanced performance across the two macroscopic alignment metrics in this experiment.}

\subsection{Timed discrepancy detection threshold study}
In the timed discrepancy detection and adjustment process, we evaluate two discrepancy metrics against real traffic data. One is the Hellinger distance of speed distributions over the past 5 seconds. The other is the MAPE of current-frame mean spacing between background vehicle pairs. If either metric exceeds a predefined threshold, the error is considered significant. If such an error is detected, the REARL method intervenes to adjust the control strategy. When REARL does not intervene, the control strategy follows that of the Base method.

\begin{figure*}[htbp]
\centering
\includegraphics[width=1.0\textwidth]{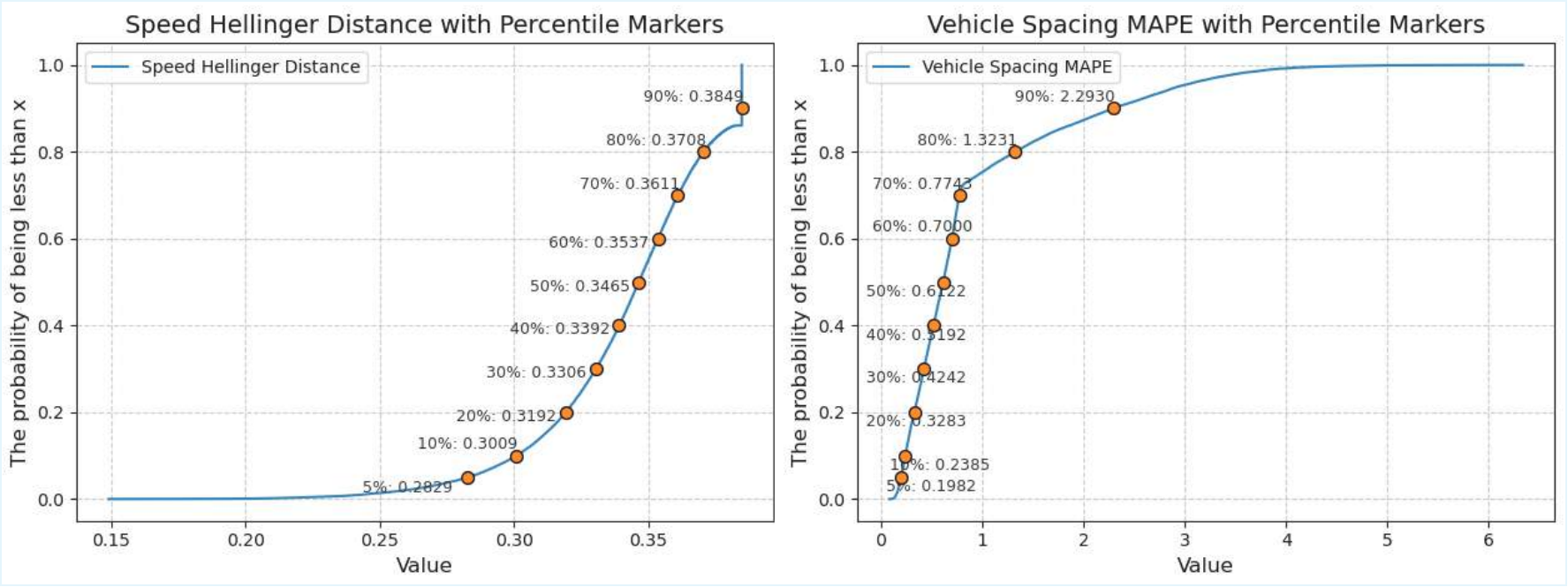}
\caption{Cumulative distribution function and percentile markers. (a) The cumulative density function (CDF) curve for the Hellinger distance of the collected speed distribution. The orange points on the curve represent the percentiles of the distribution, with the 5\%, 10\%, 20\%, 30\%, 40\%, 50\%, 60\%, 70\%, 80\%, and 90\% percentiles labeled on the plot. (b) The cumulative density function (CDF) curve for the MAPE of the collected mean spacing between pairs of vehicles distribution.}
\label{fig7}
\end{figure*}

\rev{To reduce possible overlap between threshold estimation and evaluation,} a total of 741 scenarios were divided following a 2:8 ratio. To determine the threshold, we repeated the Base method's experiment 5 times and compared it with 150 randomly selected scenarios. A total of 42,000 \rev{Hellinger} distance differences for speed distribution and 45,000 mean spacing differences between background vehicle pairs were obtained. The distribution of statistical quantiles is illustrated in Fig. \ref{fig7}. From the remaining 591 scenarios, 15 were randomly chosen for \rev{threshold hyperparameter testing without reusing the threshold-estimation subset}.

\begin{table*}[!t]
\caption{Comparison of different intervention threshold combinations}
\label{table:3}
\begin{adjustwidth}{-\dimexpr 0.15in + \oddsidemargin + \hoffset \relax}{0pt}
\noindent
\begin{tabular}{@{}
  >{\centering\arraybackslash}p{4cm}
  >{\centering\arraybackslash}p{4cm}
  >{\centering\arraybackslash}p{4cm}
  >{\centering\arraybackslash}p{4cm}
@{}}
\toprule
\raggedright \textbf{Hellinger distance threshold for speed distribution} &
\textbf{MAPE threshold for mean spacing between pairs of vehicles} &
\textbf{speed distribution (Hellinger distance)} &
\textbf{mean spacing between pairs of vehicles (MAPE)} \\
\midrule
0.2829(5\%) & 0.1982(5\%) & 0.3207 & 0.9401 \\
0.3009(10\%) & 0.1982(5\%) & 0.3471 & 1.0918 \\
0.3192(20\%) & 0.1982(5\%) & 0.3446 & 1.0294 \\
0.3306(30\%) & 0.1982(5\%) & 0.3098 & 0.8188 \\
0.2829(5\%) & 0.2385(10\%) & 0.3336 & 0.9212 \\
0.3009(10\%) & 0.2385(10\%) & 0.3223 & 0.9786 \\
0.3192(20\%) & 0.2385(10\%) & 0.3315 & 0.9029 \\
0.3306(30\%) & 0.2385(10\%) & 0.3230 & 1.3504 \\
0.2829(5\%) & 0.3283(20\%) & 0.3062 & 1.0992 \\
0.3009(10\%) & 0.3283(20\%) & 0.3299 & 1.0404 \\
0.3192(20\%) & 0.3283(20\%) & 0.3241 & 0.8160 \\
0.3306(30\%) & 0.3283(20\%) & 0.3271 & 1.2365 \\
0.2829(5\%) & 0.4242(30\%) & 0.3289 & 1.2995 \\
0.3009(10\%) & 0.4242(30\%) & 0.3335 & 1.1469 \\
0.3192(20\%) & 0.4242(30\%) & 0.2969 & 1.0534 \\
0.3306(30\%) & 0.4242(30\%) & 0.3097 & 1.0604 \\
\bottomrule
\end{tabular}
\end{adjustwidth}

\end{table*}

Based on this distribution, we selected the 5\%, 10\%, 20\% and 30\% percentiles as intervention thresholds for the REARL method, resulting in a total of 16 threshold combinations. We then conducted experiments on 15 randomly selected scenarios to determine the optimal intervention threshold combination. The experimental results in Table \ref{table:3} \rev{show how different threshold settings influence the evaluated metrics}. It is evident that a 30\% threshold for speed distribution and a 5\% threshold for mean spacing between vehicle pairs yield the best overall performance. These findings \rev{indicate that threshold selection influences the trade-off between intervention frequency and distributional alignment in the tested scenarios}.

\subsection{Trajectory clustering parameters study of the KIcluster algorithm}
To further improve the clustering performance, we employed a Bayesian optimization approach to tune the parameters of the KIcluster algorithm. Given the challenges in selecting optimal parameters for clustering algorithms, we leveraged Bayesian optimization \cite{48} to search for the best configuration in a data-driven and unbiased manner. Specifically, the optimization process utilized a Gaussian prior function and the gp\_hedge adaptive acquisition function selection strategy. The optimization ran for 500 iterations, randomly sampling 5\% of the vehicle trajectories to determine the optimal parameters. The optimal parameters for the KIcluster algorithm in this task were found to be clustering into 3 groups with 3 iterations.

\begin{figure}[!t]
\centering
\includegraphics[width=0.5\textwidth]{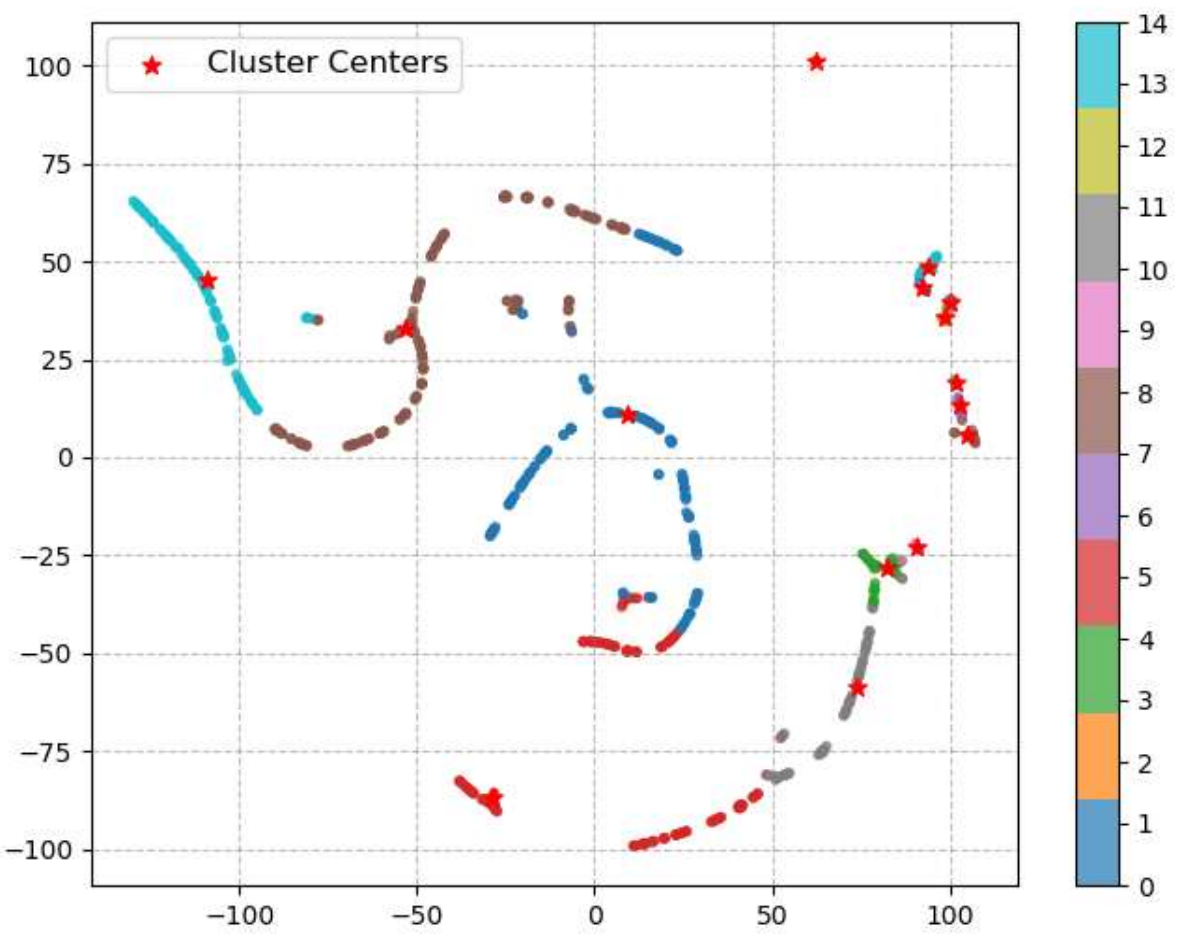}
\caption{The scenario clustering results after t-SNE dimensionality reduction. The results of the scenario clustering are visualized using t-SNE (t-Distributed Stochastic Neighbor Embedding) for dimensionality reduction, with the data points representing different vehicle trajectories across the clustering process. The color-coded data points correspond to different clusters identified by the KIcluster algorithm, with each cluster being grouped according to its statistical characteristics. The red stars represent the cluster centers, which act as the representative points of each cluster.}
\label{fig8}
\end{figure}

Based on the above optimization, the scenario clustering results are shown in Fig. \ref{fig8}. From the clustering plot, it is evident that the real traffic scenarios are effectively grouped into 15 distinct clusters, as indicated by the separation in the plot. Each grouping represents a typical traffic pattern, which is crucial for adjusting the simulation environment. The distinct separation between clusters represents the variability in driving behaviors, such as speed and spacing, that is captured from real-world data.

The linear distribution observed in the clustering plot arises because the smaller traffic scenarios are cropped from long-term traffic data. Traffic flow is typically smooth and continuous, with vehicle states gradually changing over time, such as transitioning from free-flowing to congested conditions. This state change is not abrupt, but rather gradual. Additionally, the movement of vehicles is influenced by inertia, which results in small state changes between adjacent time steps, creating continuity. t-SNE preserves these local similarities during the dimensionality reduction process, leading to the formation of a streamline-like cluster structure in the low-dimensional space, reflecting the natural transitions between different traffic states. Therefore, this linear distribution illustrates the temporal continuity and gradual state changes in traffic flow.

Moreover, the presence of outliers in the top-right corner of the plot may represent rare traffic scenarios. While these rare scenarios are infrequent, they can often contribute to safety incidents in real traffic conditions. These incidents are typically caused by the inability to handle such uncommon situations effectively. Therefore, we have retained the outliers in our analysis \rev{to preserve these rare cases in the threshold analysis, while systematic evaluation of long-tail safety-critical scenarios remains future work}.

\subsection{Inference Time Comparison}
To examine computational efficiency, we compare the inference time of three decision-making methods: highway-env's built-in rule-based method (IDM+MOBIL), PPO \cite{55}, and the proposed REARL framework. Under the same experimental setting, we record the inference time over 10 consecutive decision steps for each method and report the average, as summarized in Table \ref{table:4}.

\begin{table}[H]
\caption{Comparison of average inference time over 10 decision steps for different methods.}
\label{table:4}
\centering
\begin{tabular}{lc}
\hline
Method & Average inference time (ms) \\
\hline
IDM+MOBIL & 0.461 \\
PPO & 15.898 \\
REARL & 86241.309 \\
\hline
\end{tabular}
\end{table}

Table \ref{table:4} shows that IDM+MOBIL has the shortest inference time, averaging 0.461 ms, which confirms the efficiency of rule-based decision-making. PPO takes 15.898 ms on average, which is longer than IDM+MOBIL but still relatively efficient. By contrast, REARL requires 86241.309 ms on average for 10 consecutive decisions. This much higher cost mainly comes from the additional computation associated with discrepancy-aware detection and LLM-based reasoning in the decision process.

Overall, the results reflect a trade-off between computational efficiency and simulation fidelity. Although REARL takes much longer than IDM+MOBIL and PPO, its goal is not to serve as a lightweight \rev{vehicle-level control policy}, but to improve the realism of the simulation environment. From this perspective, \rev{the additional computational cost may be acceptable in simulation-oriented settings where fidelity is prioritized over inference efficiency}.

\section{Conclusion and future work}
In this paper, we proposed REARL, a \rev{closed-loop simulation enhancement framework} that incorporates real-world traffic data and large language models to improve the realism of autonomous driving simulation. The framework combines representative scenario construction from real traffic trajectories, discrepancy-aware online detection, and LLM-guided behavior adjustment for background vehicles. By tracking the gap between simulated traffic flow and real traffic data in terms of speed distribution, mean spacing between vehicle pairs, time headway, and lane change rate, \rev{REARL can intervene during rollout to improve the alignment between simulated traffic flow and real-world traffic patterns under the evaluated setting.} Experimental results on the HighD dataset and the highway-env platform show that REARL \rev{improves speed-distribution alignment across the tested backbones and provides the most balanced overall performance with the default Qwen3 32B configuration, while the spacing results reveal backbone-dependent trade-offs}. Among them, the default REARL with Qwen3 32B \rev{provides the most balanced performance under the evaluated metrics}, with a Hellinger distance of 0.3067, a MAPE of 0.8371, a THW of 22.8575, and a lane change rate of 0.0708. \rev{The inclusion of PPO as a stronger learning-based baseline further suggests that the observed performance gain is associated with the proposed framework design rather than solely with a particular backbone model or the use of learning-based methods.}
              
\rev{The current validation is limited to selected HighD-derived highway scenarios with a fixed four-lane highway configuration, 10 controlled vehicles, and a 60-step horizon, and therefore does not by itself demonstrate generalization to urban, mixed-traffic, or larger-scale simulation settings.} Despite these improvements, the current framework still incurs a relatively high inference cost because it relies on discrepancy detection and LLM-based reasoning. Future work will therefore focus on improving computational efficiency through lightweight decision modules, more selective intervention strategies, and faster backbone models. We also plan to \rev{systematically evaluate and extend} REARL to a wider range of traffic scenarios and rare edge cases, and to evaluate it in larger-scale and higher-fidelity simulation environments, including urban and 3D driving settings. Another direction is to further enrich the evaluation framework with additional realism-oriented metrics, such as collision rate, jerk-based comfort measures, and long-tail safety-critical event coverage, to provide \rev{a more fine-grained yet still bounded assessment of simulation fidelity}.

\bibliographystyle{IEEEtran}
\bibliography{references}
\end{document}